\documentclass[conference]{style/IEEEtran}
\usepackage{times}
\usepackage[hang,flushmargin]{footmisc} 

\usepackage[numbers]{natbib}
\usepackage{multicol}
\usepackage[bookmarks=true]{hyperref}
\hypersetup{colorlinks,breaklinks,
linkcolor=[rgb]{0.5,0.,0.},
citecolor=[rgb]{0.000,0.427,0.173},
urlcolor=[rgb]{0.031,0.318,0.612}}

\newcommand\blfootnote[1]{%
  \begingroup
  \renewcommand\thefootnote{}\footnote{#1}%
  \addtocounter{footnote}{-1}%
  \endgroup
}

\usepackage[printonlyused,withpage,nolist,nohyperlinks]{acronym}
\begin{acronym}

\acro{NMPC}{Nonlinear Model Predictive Control}
\acro{MPC}{Model Predictive Controller}
\acro{MAV}{Micro Aerial Vehicle}
\acro{DoF}{Degree of Freedom}
\acro{CoM}{Centre of Mass}
\acrodefplural{DoF}[DoFs]{Degrees of Freedom}
\acro{QP}{Quadratic Program}
\acro{ICP}{Iterative Closest Point}

\end{acronym}

\usepackage{graphics} 
\usepackage{epsfig} 
\usepackage{stfloats}
\usepackage{amsmath} 
\usepackage{amssymb}  
\usepackage{siunitx}
\usepackage{multirow}
\usepackage{siunitx}
\usepackage{url} 
\usepackage{xfrac}

\usepackage{xifthen}
\usepackage{xparse}
\usepackage{subdepth}
\usepackage{leftidx}
\usepackage{bm}
\usepackage{amsmath}

\newcommand{\bbm}{\begin{bmatrix}}
\newcommand{\ebm}{\end{bmatrix}}

\DeclareMathAlphabet{\mbf}{OT1}{ptm}{b}{n}
\newcommand{\mbs}[1]{{\bm{#1}}}

\newcommand{\mbsbar}[1]{{\overline{\boldsymbol{#1}}}}
\newcommand{\mbshat}[1]{{\hat{\boldsymbol{#1}}}}
\newcommand{\mbstilde}[1]{{\tilde{\boldsymbol{#1}}}}

\newcommand{\mbsdot}[1]{{\dot {\boldsymbol{#1}}}}

\newcommand{\mbfbar}[1]{{\overline{\mbf{#1}}}}
\newcommand{\mbfhat}[1]{{\hat{\mbf{#1}}}}
\newcommand{\mbftilde}[1]{{\tilde{\mbf{#1}}}}

\newcommand{\mbfdot}[1]{{\dot{\mbf{#1}}}}

\newcommand{\cframe}[1]{{\smash{\protect\underrightarrow{\mathcal{F}}_{#1}}}}

\DeclareMathAlphabet{\mathbfit}{OML}{cmm}{b}{it}
\newcommand{\homo}[1]{{\mathbfit{#1}}}

\newcommand{\mbfh}[1]{{\homo{#1}}}

\newcommand{\argmin}{\operatornamewithlimits{argmin}}

\newcommand{\pos}[2]{\leftidx{_{#1}}{ \mbf r}{_{#2}}} 
\newcommand{\posdot}[2]{\leftidx{_{#1}}{\mbfdot r}{_{#2}}} 

\newcommand{\vel}[3]{\leftidx{_{#1}}{\mbf v}{\IfValueTF{#2}{_{#2#3\hspace{2pt}}}{}}} 
\newcommand{\veltilde}[3]{\leftidx{_{#1}}{\mbftilde v}{\IfValueTF{#2}{_{#2#3\hspace{2pt}}}{}}} 
\newcommand{\velbar}[3]{\leftidx{_{#1}}{\mbfbar v}{\IfValueTF{#2}{_{#2#3\hspace{2pt}}}{}}} 
\newcommand{\velhat}[3]{\leftidx{_{#1}}{\mbfhat v}{\IfValueTF{#2}{_{#2#3\hspace{2pt}}}{}}} 
\newcommand{\veldot}[3]{\leftidx{_{#1}}{\mbfdot v}{\IfValueTF{#2}{_{#2#3\hspace{2pt}}}{}}} 

\newcommand{\myvec}[2]{\leftidx{_{#2}\hspace{-1pt}}{\mbf #1}{}} 

\newcommand{\acc}[3]{\leftidx{_{#1}}{\mbf a}{\IfValueTF{#2}{_{#2#3\hspace{2pt}}}{}}} 
\newcommand{\acctilde}[3]{\leftidx{_{#1}}{\mbftilde a}{\IfValueTF{#2}{_{#2#3\hspace{2pt}}}{}}} 
\newcommand{\accbar}[3]{\leftidx{_{#1}}{\mbfbar a}{\IfValueTF{#2}{_{#2#3\hspace{2pt}}}{}}} 

\newcommand{\rotvel}[3]{\leftidx{_{#1}}{\mbs \omega}{\IfValueTF{#2}{_{#2#3\hspace{2pt}}}{}}} 
\newcommand{\rotveltilde}[3]{\leftidx{_{#1}}{\mbstilde \omega}{\IfValueTF{#2}{_{#2#3\hspace{2pt}}}{}}} 
\newcommand{\rotvelbar}[3]{\leftidx{_{#1}}{\mbsbar \omega}{\IfValueTF{#2}{_{#2#3\hspace{2pt}}}{}}} 
\newcommand{\rotvelhat}[3]{\leftidx{_{#1}}{\mbshat \omega}{\IfValueTF{#2}{_{#2#3\hspace{2pt}}}{}}} 
\newcommand{\rotveldot}[3]{\leftidx{_{#1}}{\mbsdot \omega}{\IfValueTF{#2}{_{#2#3\hspace{2pt}}}{}}} 

\newcommand{\C}[2]{\leftidx{}{\mbf C}{_{#1#2\hspace{2pt}}}} 
\newcommand{\T}[2]{\leftidx{}{\mbfh T}{_{#1#2\hspace{2pt}}}} 
\newcommand{\q}[2]{\leftidx{}{\mbf q}{_{#1#2\hspace{2pt}}}} 
\newcommand{\qdot}[2]{\leftidx{}{\mbfdot q}{_{#1#2\hspace{2pt}}}} 

\begin{document}

\title{Aerial Manipulation Using Hybrid Force and Position NMPC Applied to Aerial Writing}
\author{\IEEEauthorblockN{Dimos Tzoumanikas\IEEEauthorrefmark{1}, Felix Graule\IEEEauthorrefmark{1}\IEEEauthorrefmark{2}, Qingyue Yan\IEEEauthorrefmark{1}, Dhruv Shah\IEEEauthorrefmark{3}, Marija Popovi\'{c}\IEEEauthorrefmark{1} and Stefan Leutenegger\IEEEauthorrefmark{1}}
 \IEEEauthorblockA{%
    \IEEEauthorrefmark{1}Imperial College London,
    \IEEEauthorrefmark{2}ETH Zurich,
    \IEEEauthorrefmark{3}University of California, Berkeley
  }}
\maketitle
\begin{abstract}
Aerial manipulation aims at combining the maneuverability of aerial vehicles with the manipulation capabilities of robotic arms. This, however, comes at the cost of the additional control complexity due to the coupling of the dynamics of the two systems. In this paper we present a \ac{NMPC} specifically designed for \acp{MAV} equipped with a robotic arm. We formulate a hybrid control model for the combined MAV-arm system which incorporates interaction forces acting on the end effector. We explain the practical implementation of our algorithm and show extensive experimental results of our custom built system performing multiple `aerial-writing' tasks on a whiteboard, revealing accuracy in the order of millimetres.
\end{abstract}

\IEEEpeerreviewmaketitle
\blfootnote{This work has been supported by the EPSRC grant Aerial ABM EP/N018494/1 and Imperial College London.\vspace{0.02em}}
\blfootnote{Corresponding author email: \texttt{dt214@ic.ac.uk}\vspace{0.02em}}
\blfootnote{Video of the experiments: \url{https://youtu.be/iE--MO0YF0o}}
\section{Introduction}
\label{sec:introduction}
Over the past decades, aerial manipulation has received great attention in the robotics research community, with many different systems in use \cite{Ruggiero2018, Bonyankhamseh2018}. The tasks solved by aerial manipulators range from grasping, fetching, and transporting arbitrary objects to pushing against fixed surfaces. Potential use cases are numerous: inspection of infrastructure like bridges or manufacturing plants \cite{Ikeda2017, Trujillo2019, Bodie2019}, physical interaction through tools like grinding, welding, drilling and other maintenance work in hard-to-reach places \cite{Papachristos2014, Bodie2019}, and the autonomous pick-up and transport of objects \cite{Kessens2016, Augugliaro2014}. All of these tasks require the \ac{MAV} to be equipped with an additional mechanism which we refer to as the end effector. The coupling of the \ac{MAV} dynamics with the moving end effector poses an interesting challenge from a control perspective given the inherent instability of \acp{MAV}. 

The requirements for high precision in real world aerial manipulation applications further increases the difficulty of the task. Efforts made into this direction include \cite{Darivianakis2014,Bodie2019}, where the authors use a fixed end effector on a underactuated and an omnidirectional \ac{MAV} respectively. Compared to the former method, our approach achieves higher accuracy and flexibility in terms of potential use cases due to our moving end effector. In comparison to the second approach, we achieve on par precision while relying on a simpler, underactuated platform. In summary, we claim to show the following contributions: 
\begin{itemize}
    \item We present a hybrid model which captures the non-linear dynamics of the MAV and considers the quasi-static forces introduced by the attached manipulator.
    \item We use this generic model in an \ac{NMPC} jointly controlling the MAV and arm motion.
    \item We experimentally evaluate our method in `aerial-writing' tasks using our custom built system. Our results demonstrate high repeatability and accuracy in the order of millimetres across multiple trajectories of varying difficulty.  
\end{itemize}

This paper is organised as follows: In Section \ref{sec:related_work} we give an overview of the related work on aerial manipulation. In Sections \ref{sec:notation} and \ref{sec:system_overview} we describe our notation, geometric arrangement, and the software architecture of our system. We explain the method in detail in Section \ref{sec:method} followed by the experimental results in Section \ref{sec:experiments}. Finally, in Section \ref{sec:discussion} we discuss our findings and conclude in Section \ref{sec:conclusion}.
\begin{figure}[t]
\centering
  \includegraphics[width=0.48\textwidth]{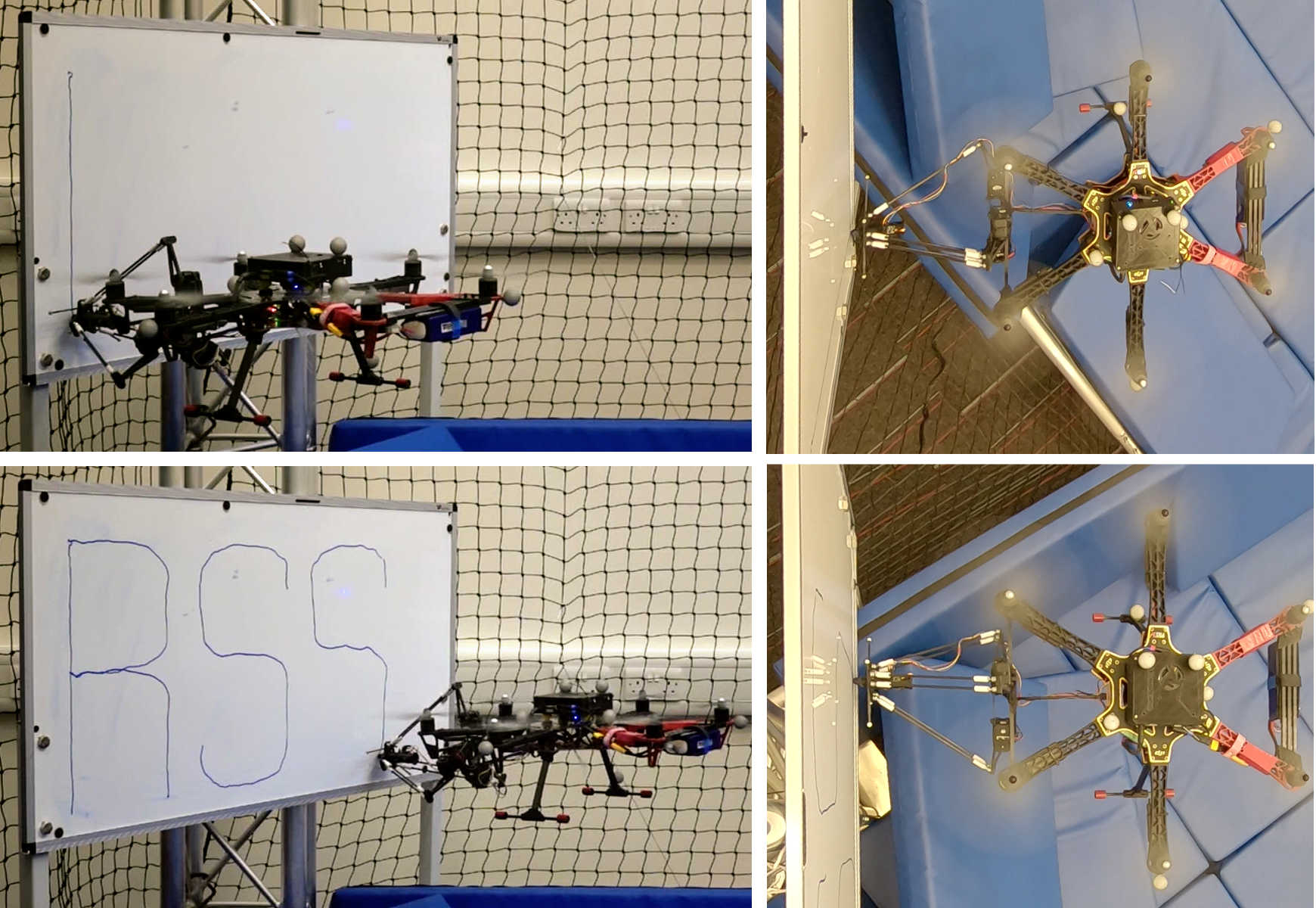}
\caption{Our MAV-arm system performing an `aerial-writing' task.}
\label{fig:first_figure}
\end{figure}
 
\section{Related Work}
\label{sec:related_work}
Aerial manipulation systems can be broadly distinguished based on the \ac{MAV} type (as being omnidirectional or underactuated) and the end effector (as being fixed or moving). In general, using an omnidirectional \ac{MAV} to fulfill complex aerial tasks does not require a moving end effector as the necessary 6 \acp{DoF} are provided by the \ac{MAV} itself. Examples include the works presented by \citep{Brescianini2016, Brescianini2018, Ryll2019}. \citet{Brescianini2016} show an omnidirectional \ac{MAV} that achieves 6-\ac{DoF} motion by using eight fixed rotors in a non-co-planar configuration. In a subsequent study \citep{Brescianini2018}, the same platform is used with a fixed end effector to fetch moving objects. Using a similar approach with a fixed configuration of six tilted rotors, \citet{Ryll2019} propose a novel paradigm to control all 6 \acp{DoF} of the \ac{MAV} while using a rigid end effector to exert forces and torques independently. The system is demonstrated in numerous experimental tasks that include contact, e.g.\ surface sliding and tilted peg-in-hole task. 

In a different approach \cite{Kamel2018}, a setup consisting of six rotors which actively control the thrust direction is proposed. \citet{Bodie2019} leverage this system to solve a variety of aerial manipulation tasks with a rigidly mounted, low complexity end effector. The authors further show precise force control when in contact with unstructured environments while relying on visual-inertial state estimation. While this platform allows for accurate 6-\ac{DoF} flight and longitudinal force exertion with a relatively simple control method, it is mechanically more complex and thus more costly compared to classical multicopter platforms. Recently, a similar approach was followed by \citet{Trujillo2019}, who introduce AeroX, an omnidirectional octacopter for contact-based inspection. Their end effector design minimises the torque caused by contact forces and features wheels on its base to allow moving along a surface while remaining in contact. In \cite{Papachristos2014, Papachristos2014ICRA} the authors show a less complex but highly capable tri-tilt-rotor \ac{MAV} for surface grinding and obstacle manipulation. The control model consists of two disjoint modes: one for free-flight and another for physical interaction.

Employing an underactuated \ac{MAV} to perform aerial manipulation typically increases the complexity of the end effector since the latter has to provide the additional \acp{DoF}. To investigate this, many different end effector designs have been proposed over the last years. We categorise these works by the increasing complexity of the end effector. In \cite{Darivianakis2014} an underactuated \ac{MAV} with a fixed end effector was used for performing contact-based tasks. The authors use a switching mode linear \ac{MPC} with different control models for free flight and in-contact operation. Similarly to our work, they benchmark their approach by performing `aerial-writing'. In \citep{Mellinger2011}, the authors use different lightweight, low complexity grippers to perch, pick up, and transport payload. Using a least squares approach, they estimate the inertial parameters of the grasped objects and use this information to adapt the controller and improve tracking. 
Moving up in terms of complexity, \citet{Kim2013} suggest mounting a 2-\ac{DoF} robotic arm on a
\ac{MAV} to allow grasping and transporting of objects. The authors propose an adaptive sliding mode controller for the combined system. In \cite{Orsag2014} the authors present an aerial manipulator with two robotic 2-\ac{DoF} arms to open a valve. The \ac{MAV} and arms are controlled as a coupled system which is modelled as a switched nonlinear system during valve turning. In an approach very similar to ours, \citet{Lunni2017} propose an \ac{NMPC} which jointly controls the motion of an underactuated \ac{MAV} and the attached manipulator. However, the control model is only formulated for free flight operation and the presented experiments do not include contact. In a more recent work, \citet{Suarez2018} propose a lightweight, human-sized dual arm system designed to minimise the inertia transferred to the \ac{MAV}. Each of the two arms add 5 \acp{DoF} to the system and the arm control law applied takes into account that low-cost servo motors do not allow torque control but require position commands. Further, a torque estimator is used to predict the torques produced by the servos and inform the \ac{MAV} control algorithm accordingly. In order to minimise such disturbances coming from the end effector, \citet{Nayak2018} propose a lightweight design mounted on top of an \ac{MAV}. While attaching a serial robotic arm on an MAV increases the number of tasks it can perform, they only provide limited precision when using low cost and lightweight actuators. Some previous efforts try to mitigate this by mounting a parallel delta arm on an \ac{MAV} instead. \cite{Kamel2016_1, Kamel2016_2} demonstrate a multi-objective dynamic controller which also considers dynamic effects between the \ac{MAV} and its 3-\ac{DoF} delta arm. 
The same system is used in \cite{Steich2016} to inspect tree cavities with a camera mounted at the end effector.

In our work we use a low mechanical complexity setup consisting of an underactuated hexacopter and a 3-\ac{DoF} parallel arm. We propose a general method which could be transferred to other aerial manipulation platforms. We extend the \ac{NMPC} framework presented in \cite{Tzoumanikas2020} for manipulation tasks and use a hybrid model that captures the effect of contact and coupled \ac{MAV}-arm dynamics. Our main focus is on the developed software while we use the presented hardware platform to showcase our method. To our best knowledge, we achieve unprecedented accuracy in experiments requiring contact by using an underactuated \ac{MAV}-delta arm system.

\section{Notation and Coordinate Frames}
\label{sec:notation}
We denote vectors as bold lower case symbols, e.g.\ $\myvec{v}{}$. We use left-hand subscripts, e.g.\ $\myvec{v}{A}$, to indicate the coordinate representation in the $\cframe{A}$ frame of reference. The rotation matrix $\C{A}{B}$ changes the representation of the vector $\myvec{v}{B}$ from $\cframe{B}$ to $\cframe{A}$ as $\myvec{v}{A}=\C{A}{B}\myvec{v}{B}$. Analogously to the rotation matrix $\C{A}{B}$, we use the quaternion $\q{A}{B}$ with $\otimes$ denoting the quaternion multiplication. 
We use $[\myvec{v}{}]^{\times}$ to denote the skew symmetric matrix of the vector $\myvec{v}{}$. The motion of the \ac{MAV} body frame $\cframe{B}$ (\emph{x}: forward, \emph{y}: left, \emph{z}: upward) is expressed with respect to the World frame $\cframe{W}$ (\emph{z}: upward). 
\section{System Overview}
\label{sec:system_overview}
\begin{figure}[htb]
    \centering
    \includegraphics[width=0.48\textwidth]{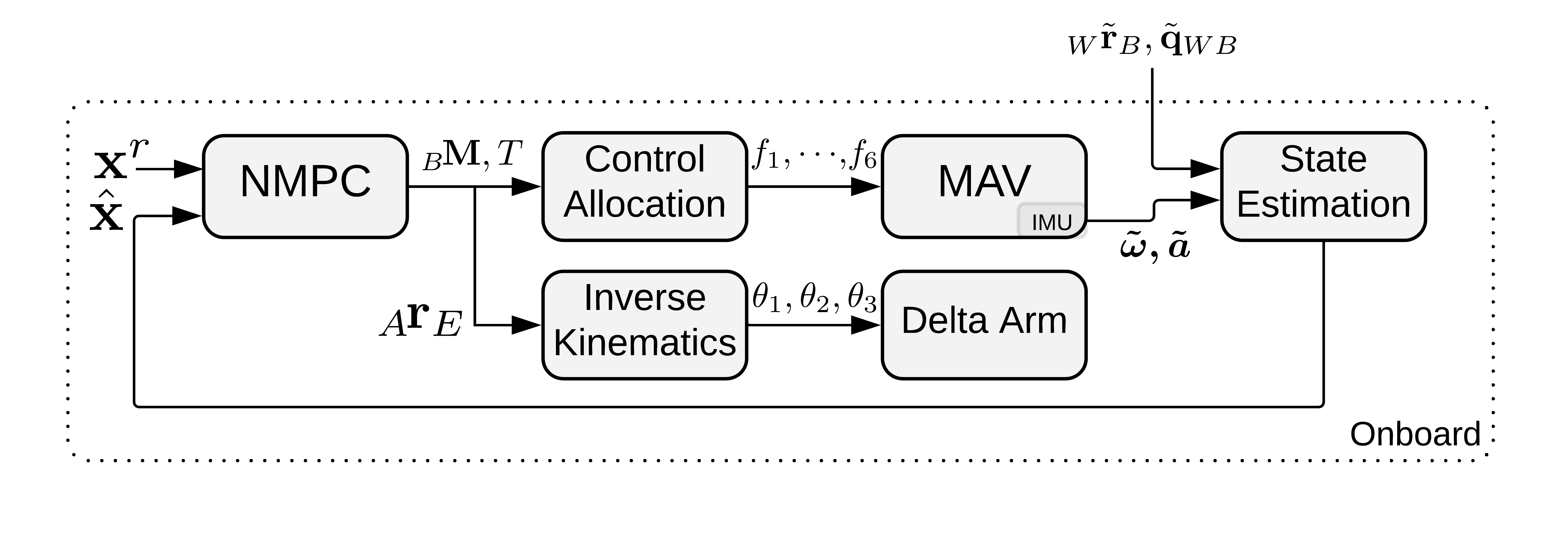}
    \vspace{-1em}
    \caption{An overview of the software running onboard our \ac{MAV} in an aerial manipulation task. We use ROS to interface with the \ac{MAV} and the motion capture system while the other software blocks form a single executable.}
    \label{fig:system_overview}
    \end{figure}

\begin{figure}[htb]
    \centering
    \includegraphics[width=0.48\textwidth]{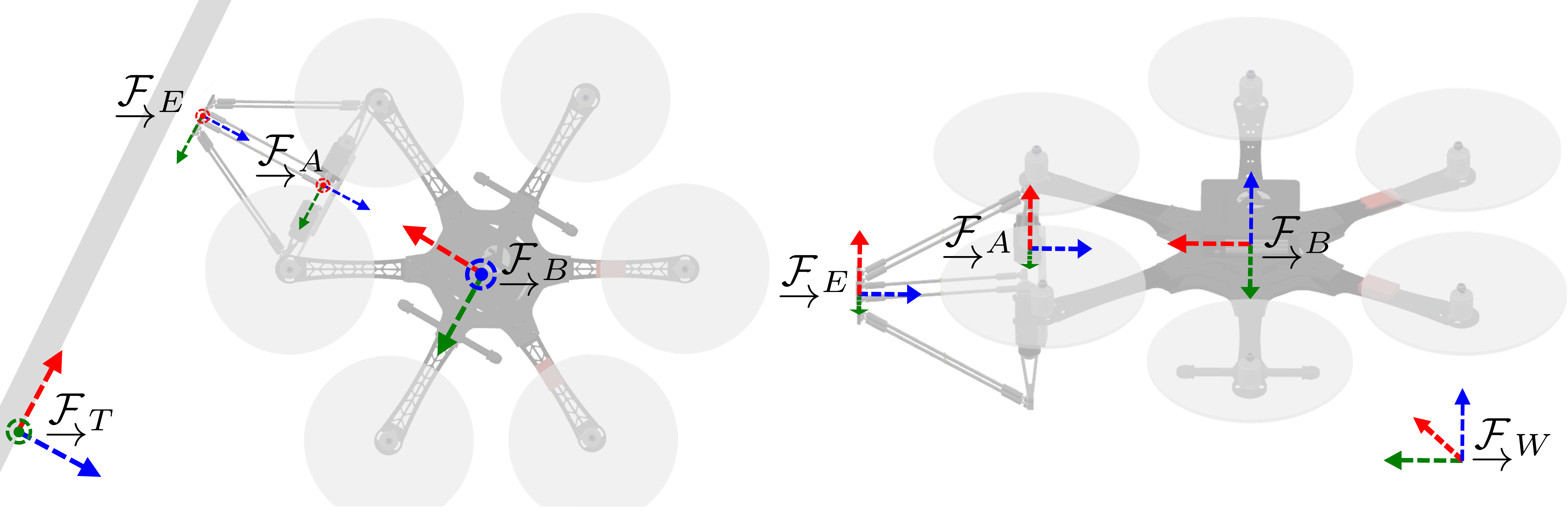}
    \caption{The coordinate frames used in this paper. Specifically, $\cframe{W}$, $\cframe{B}$, $\cframe{A}$, $\cframe{E}$, and $\cframe{T}$ stand for the World, \ac{MAV} body, arm, end effector, and contact frame, respectively.}
    \label{fig:coordinate_frames}
\end{figure}

An overview of the different software components of the proposed system is outlined in Figure \ref{fig:system_overview}. The \ac{NMPC} is given full state trajectory commands for the \ac{MAV} and the end effector corresponding to a given aerial manipulation task. Based on those references and the estimated system state, it produces the desired \ac{MAV} body moments, collective thrust, and end effector position. The control allocation block is responsible for converting the moments and thrust into individual motor commands while the inverse kinematics block computes the desired link angles for the given end effector position. All algorithms run onboard at a rate of 100~\si{\hertz} while the estimate of the \ac{MAV} position and orientation is provided externally. The different coordinate frames are displayed in Figure \ref{fig:coordinate_frames}. 
\section{Method}
\label{sec:method}
\subsection{Hybrid Modelling}
The standard Newton-Euler equations are used to model the combined MAV-arm dynamics. We model the \ac{MAV} as a single rigid body object and only consider quasi-static forces introduced by the arm dynamics and its interaction with the environment. 
Overall, the combined dynamics take the following form:
\begin{subequations}
\label{eq:mav_dynamics} 
\begin{align}
 \posdot{W}{B} &= \vel{W}{B}{}, \\ 
 \qdot{W}{B}   &= \frac{1}{2} \mathbf{\Omega}(\rotvel{B}{}{})\q{W}{B}, \\
 \veldot{W}{B}{} &= \frac{1}{m_{c}} \C{W}{B}(\myvec{F_{r}}{B} + \myvec{F_{e}}{B}) +\myvec{g}{W},  \\
 \rotveldot{B}{}{} &= \mathbf{J_c}^{-1} (\myvec{M_{r}}{B} + \myvec{M_{e}}{B}- \rotvel{B}{}{} \times \mathbf{J_c} \rotvel{B}{}{}), \\
  \mathbf{\Omega}(\rotvel{B}{}{}) &= \begin{bmatrix}
                                    \rotvel{B}{}{}^{\times} &  \rotvel{B}{}{} \\ 
                                    -\rotvel{B}{}{}^{\top}   &      0
                                   \end{bmatrix},
\end{align}
\end{subequations}
where $m_c$, $\mathbf{J_c}$ are the combined \ac{MAV}-arm mass and inertia tensors, respectively and $\myvec{g}{W}$ the acceleration due to gravity. Regarding the forces and moments $\myvec{F_i}{B}, \myvec{M_i}{B}$, we use the subscript $i \in \{r,e\}$ to distinguish the ones generated by the \ac{MAV} motors $r$ from the ones caused by the end effector movement  $e$ and its potential contact with the environment. In our system, the \ac{MAV} motor-generated forces and moments are given by: 
\begin{subequations}
\label{eq:motor_forces_moments}
\begin{align}
    \myvec{F_{r}}{B} &:= \begin{bmatrix}0,0,T\end{bmatrix}^{\top},~ T = \sum\limits_{i=1}^{6}f_i,  \\
    \myvec{M_{r}}{B} &:= \sum\limits_{i=1}^{6} \Big( f_i \myvec{r}{B}_i\times \myvec{e}{B}_z
                + (-1)^{i+1}k_m f_i\myvec{e}{B}_z\Big),
                                \end{align}
\end{subequations}
with $f_i\in \mathbb{R}$ the thrust produced by the $i^{\text{th}}$ motor, $\myvec{r}{B}_i$ its position with respect to $\cframe{B}$, $k_m$ the known thrust to moment coefficient and $\myvec{e}{B}_z=[0, 0, 1]^{\top}$.
Equation \eqref{eq:motor_forces_moments} can be summarised as $\begin{bmatrix}\myvec{M_{r}}{B}^{\top}, & T \end{bmatrix}^{\top} = \mathbf{A}\begin{bmatrix}
                                    f_1 & f_2 & \hdots & f_6
                                \end{bmatrix}^{\top}$ with $\mathbf{A}\in\mathbb{R}^{4\times 6}$ the allocation matrix related to the \ac{MAV} geometry as described in  \cite{Tzoumanikas2020}.
$\myvec{F_e}{B}$ and $\myvec{M_e}{B}$ are given by:
\begin{subequations}
\label{eq:ee_forces_moments}
\begin{align}
    \myvec{F_e}{B} &:= \C{B}{E} \myvec{F_c}{E},\\
    \myvec{M_e}{B} &:= \pos{B}{E}\times\myvec{F_e}{B} + (\pos{B}{E}-\pos{B}{E_0})\times(\C{B}{W} m_e~\myvec{g}{W}), \label{eq:ee_moment}
                                \end{align}
\end{subequations}
where $\myvec{F_c}{E}$ is the contact force acting on the end effector expressed in its frame  $\cframe{E}$ and $\pos{B}{E_0}\in\mathbb{R}^3$ the nominal end effector position which results in no \ac{CoM} displacement. The two terms in \eqref{eq:ee_moment} represent the moments due to contact and due to the displacement of the \ac{CoM} respectively. The combined mass $m_c := m + m_e$ is the sum of the \ac{MAV} and end effector mass, respectively, while the combined rotational inertia can be computed as $\mathbf{J_c} := \mathbf{J} + m_e\text{diag}(\pos{B}{E}-\pos{B}{E_0})^{2}$ with $\mathbf{J}=\text{diag}(J_x, J_y, J_z)$ the inertia tensor of the \ac{MAV} (including the arm in nominal position) and $\text{diag}(\cdot)$ the corresponding diagonal matrix. 

Regarding the contact force, we assume that this can be approximated with a linear spring model as: 
\begin{equation}
    \myvec{F_c}{E} = \C{E}{T}(k_s \pos{T}{E_z}) ,
    \label{eq:contact_model}
\end{equation}
where $k_s$ is a known spring coefficient and $\pos{T}{E_z}$ is the normal component of the contact surface penetration. This way, the controller can anticipate contact before it even happens and there is no need for a switching mode controller (one for free flight and another one for contact dynamics). 
\subsection{Model Based Control}
For the control formulation we define the following control state and input:
\begin{subequations}
\label{eq:control_state_and_input}
\begin{align}
    \myvec{x}{} &:= \begin{bmatrix}
                  \pos{W}{B}^{\top}, \vel{W}{B}{}^{\top}, \q{W}{B}^{\top}, \rotvel{B}{}{}^{\top}
                  \end{bmatrix}^{\top} \in \mathbb{R}^6 \times S^3 \times \mathbb{R}^3, \\
    \myvec{u}{} & := \begin{bmatrix}
                    \myvec{M_{r}}{B}^{\top}, T, \pos{A}{E}^{\top}
                    \end{bmatrix}^{\top} \in \mathbb{R}^7.
\end{align}
\end{subequations}
Note that we use $\pos{B}{E}$ for the formulation of the control model, while $\pos{A}{E}$ is used in the control input. We use the constant and known homogeneous transformation $\T{B}{A}$  to change the  coordinate representation of these position vectors.

We use the following error functions for the position of the \ac{MAV}, the position of the end effector, the \ac{MAV} linear and angular velocity, the orientation, the contact force and the control input, respectively:  
\begin{subequations}
    \label{eq:error_functions}
    \begin{align}
    \myvec{e}{}_{rB} &= \pos{W}{B} - \pos{W}{B}^{r},  \\
    \myvec{e}{}_{rE} &= \pos{W}{E} - \pos{W}{E}^{r},  \\
    \myvec{e}{}_{v} &= \vel{W}{B}{} - \vel{W}{B}{}^{r},  \\
    \myvec{e}{}_{\omega} &= \rotvel{B}{}{} -  \C{B}{B^r}\rotvel{B^r}{}{}^{r},  \\
    \myvec{e}{}_{q} &= [\q{W}{B}^{-1} \otimes \q{W}{B}^{r}]_{1:3},  \\ 
    e_{f} &= f_{c} - f_{c}^{r},  \\
    \myvec{e}{}_{u} & = \myvec{u}{} - \myvec{u}{}^{r},
    \end{align}
\end{subequations}
with $f_c:=\myvec{F_{c_{z}}}{E}$ and the superscript $r$ used to denote the time-varying reference quantities. The optimal input sequence $\myvec{u}{}^{*}$ is obtained by the online solution of the following constrained optimisation problem:
\begin{subequations}
\label{eq:ocp}
\begin{align}
   \myvec{u}{}^* = \argmin\limits_{\myvec{u}{}_{0}, \ldots, \myvec{u}{}_{N_f}} &\Big\{\Phi(\myvec{x}{}_{N_f},\myvec{x}{}^r_{N_f}) + \sum\limits_{n=0}^{N_f-1}   L(\myvec{x}{}_{n},\myvec{x}{}^r_{n},\myvec{u}{}_{n}) \Big\},  \\
    \text{s.t. : } &\myvec{x}{}_{n+1} = \myvec{f}{}_{d}(\myvec{x}{}_{n},\myvec{u}{}_{n}), \\
                   &\myvec{x}{}_{0} = \hat{\myvec{x}{}},  \\
                    &{u}_\text{lb} \leq {u}_{i} \leq {u}_\text{ub}, \quad i=1,\ldots,7,
\end{align}
\end{subequations}
with $N_f$ the discrete horizon length, $\myvec{f}{}_{d}$ the discrete version of the dynamics given in Equations  \eqref{eq:mav_dynamics}~--~\eqref{eq:ee_forces_moments}, $\hat{\myvec{x}{}}$ the latest state estimate and ${u}_\text{lb}, {u}_\text{ub}$ appropriate lower and upper bounds for the control input defined in \eqref{eq:control_state_and_input}. For the intermediate $L$ and final terms $\Phi$ we use quadratic costs of the form $\myvec{e}{}_i^{\top} \mathbf{Q}_i \myvec{e}{}_i$ $\forall \myvec{e}{}_i \in \{\myvec{e}{}_{rB}, \myvec{e}{}_{rE}, \myvec{e}{}_{v},
\myvec{e}{}_{\omega}, \myvec{e}{}_{q}, e_{f}, \myvec{e}{}_{u}\}$ as defined in \eqref{eq:error_functions} where the gain matrices $\mathbf{Q}_i\succcurlyeq 0$ were experimentally tuned. 

The optimal control problem is implemented using a modified version of the CT toolbox \cite{ct} with a 10~\si{\milli\second} discretisation step and a 2~\si{\second} constant prediction horizon. We use a Runge-Kutta 4 integration scheme followed by a re-normalisation for the quaternion. As common in receding horizon control, the first input $\myvec{u}{}_{0}^{*}$ is applied to the system and the whole process is repeated once a new state estimate $\hat{\myvec{x}{}}$ becomes available. The motor commands $\myvec{f}{}=\begin{bmatrix}
                                    f_1 & f_2 & \hdots & f_6
                                \end{bmatrix}^{\top}$ for the \ac{MAV} are obtained by solving the following \ac{QP}: 
\begin{subequations}
\label{eq:control_allocation_mav}
\begin{align}
\myvec{f}{}^* = &\argmin\limits_{\myvec{f}{}} \Big( \left\Vert
 \mbf{A} \myvec{f}{} - \myvec{u}{}_{{0}_{1:4}}^*\right\Vert_{\mbf{W}}^{2} + \lambda\left\Vert \myvec{f}{} \right\Vert_{2}^{2}\Big),  \\
 \text{s.t. : } &f_\text{min} \leq f_i \leq f_\text{max},\quad i=1,\ldots,6,
\end{align}
\end{subequations}
where $f_\text{min}, f_\text{max} \in \mathbb{R}$ are the minimum and the maximum motor thrust, $\mbf{W}\in\mathbb{R}^{4\times 4}$ is a gain matrix and $\lambda=10^{-7}$ a regularisation parameter.
The end effector position commands $\myvec{u}{}_{{0}_{5:7}}^*$ are mapped into servo angle commands $\theta_1, \theta_2, \theta_3$ by solving the inverse kinematics problem for the delta arm explained in Section \ref{sec:delta_arm_kinematics}. In the case of an infeasible (e.g.\ outside the arm's workspace) or unsafe end effector position command (e.g.\ one that results collision between the \ac{MAV} propellers and the arm's links), the position command is reprojected onto the boundary of the feasible and safe to operate workspace. In practice, this was rarely the case as the \ac{MAV} and end effector reference trajectories are designed so that the end effector operates close to its nominal position. In this way the usable workspace is maximised while the effect of the \ac{CoM} displacement (which is captured by our control model) is minimised.

Our C++ implementation of the above, requires 6.7~\si{\milli\second} with a standard deviation of 0.57~\si{\milli\second} per iteration. On average, 98\% of the computation time is spent on the optimisation problems defined in \eqref{eq:ocp} and \eqref{eq:control_allocation_mav}.

We would like to highlight that our method is generic enough to be applied to different types of vehicles such as omnidirectional ones and/or other types of manipulators. In these cases, the control input, the control allocation and the arm kinematics have to be adapted based on the vehicle and manipulator type. Similarly, the model can easily be extended to capture aerodynamic friction, gyroscopic moments, handle multiple contact points, or use more sophisticated contact models (e.g.\ ones that include a combination of linear springs and dampers). Similarly, the writing task which we use for the experimental evaluation, is just an example application that requires precision. We believe that our algorithms are adaptable to other tasks such as inspection through contact.
\subsection{Delta Arm Kinematics}
\label{sec:delta_arm_kinematics}
\begin{figure}[htb]
\centering
  \includegraphics[width=0.48\textwidth]{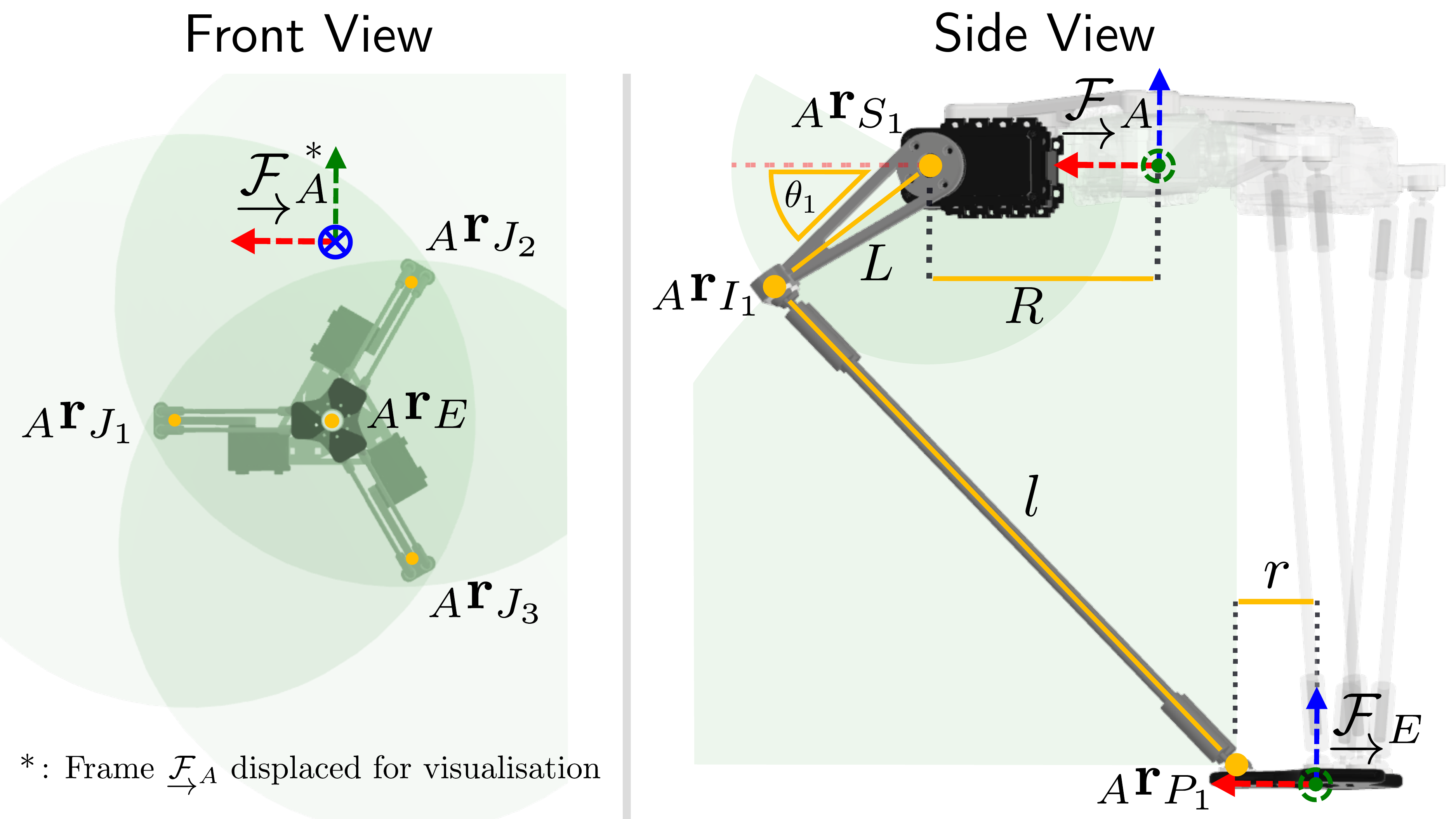}
\caption{Two different views of a 3D model of the delta arm used. The green areas show the virtual spheres and disks used for the solution of the forward and inverse kinematics.}
\label{fig:delta_arm_kinematics}
\end{figure}

Our \ac{MAV} is equipped with a custom built 3-\ac{DoF} delta arm \cite{delta_arm_paper}. Its main advantages are speed, as its few moving parts are made of lightweight materials, precision and the easy to solve forward and inverse kinematics. The forward kinematics problem (i.e.\ determining the position of the end effector $\pos{A}{E}$ given the joint angles $\theta_1, \theta_2, \theta_3$) can be solved by computing the intersection points of three spheres (shown in Figure \ref{fig:delta_arm_kinematics}) of radius $l$ with the following centres:
\begin{subequations}
\label{eq:arm_forward_kinematics_sphere_centers}
\begin{align}
\pos{A}{J_1} &= \big(R-r + L \cos(\theta_1)\big)\myvec{e_x}{A} - \sin(\theta_1)\myvec{e_z}{A}, \nonumber \\
\pos{A}{J_2} &= \mathbf{C}_z(120^o)\Big(\big(R-r + L \cos(\theta_2)\big)\myvec{e_x}{A} - \sin(\theta_2)\myvec{e_z}{A}\Big), \nonumber\\
\pos{A}{J_3} &= \mathbf{C}_z(240^o)\Big(\big(R-r + L \cos(\theta_3)\big)\myvec{e_x}{A} - \sin(\theta_3)\myvec{e_z}{A}\Big),  \nonumber
\end{align}
\end{subequations}
 where $R, r, L$ correspond to the arm physical parameters shown in Figure \ref{fig:delta_arm_kinematics}, $\myvec{e_x}{A}=[1, 0, 0]^{\top}$, $\myvec{e_z}{A}=[0, 0, 1]^{\top}$
 and $\mathbf{C}_z(120^o), \mathbf{C}_z(240^o)$ rotation matrices of $120^o$ and $240^o$ degrees around $\myvec{e_z}{A}$.  The maximum number of intersection points is two which corresponds to an end effector position above ($\pos{A}{E_z} > 0 $) and below ($\pos{A}{E_{z}} < 0 $) the arm base. In our setup, solutions above the arm base are mechanically impossible and thus rejected. For the inverse kinematics the intersection between a sphere with radius $l$ and a circular disk with radius $L$ has to be computed for every joint angle. For the first joint, as shown in Figure \ref{fig:delta_arm_kinematics}, the centre of the sphere is $\pos{A}{P_1} = \pos{A}{E} + r \myvec{e_x}{A}$ with $\myvec{e_x}{A}=[1, 0, 0]^{\top}$ while the centre of the circular disk is $\pos{A}{S_1} =  R\myvec{e_x}{A}$. Given the intersection point $\pos{A}{I_1}$, the joint angle can be recovered as $\theta_1= \arcsin(\sfrac{\pos{A}{I_{1z}}}{L})$. The joint angles $\theta_2$ and $\theta_3$ can be computed by performing the same procedure for the spheres with centres $\pos{A}{P_2}, \pos{A}{P_3}$, same radius $l$ and the unit disks centred at $\pos{A}{S_2}$ and $\pos{A}{S_3}$ with radius $L$. The points $\pos{A}{P_i}, \pos{A}{S_i}~ \forall i={2,3}$ can be easily computed as follows:
 \begin{subequations}
 \label{eq:arm_inv_kinematics_sphere_centers}
 \begin{align}
  \pos{A}{P_2} &= \pos{A}{E} +  r\ \mathbf{C}_z(120^o)\myvec{e_x}{A}, \\
  \pos{A}{P_3} &= \pos{A}{E} +  r\ \mathbf{C}_z(240^o)\myvec{e_x}{A}, \\
  \pos{A}{S_2} &= \mathbf{C}_z(120^o)\pos{A}{S_1}, \\
  \pos{A}{S_3} &= \mathbf{C}_z(240^o)\pos{A}{S_1}.
 \end{align}
 \end{subequations}
\subsection{Trajectory Generation}
\label{sec:trajectory_generation}
We use a trajectory generator to map arbitrary sets of characters to end effector trajectories. We use a constant acceleration motion model to generate trajectories with a smooth velocity profile. This is of special importance when the reference path contains sharp edges. Through our software we can adjust the velocity and acceleration profile by changing the maximum  $\|\vel{W}{E}{}^r\|$ and $\|\acc{W}{E}{}^r\|$. Once the trajectory for the end effector has been computed, we proceed with the computation of the reference position $\pos{W}{B}^{r}$ and velocity $\vel{W}{B}{}^{r}$ for the \ac{MAV} as follows: 
\begin{subequations}
\begin{align}
    \pos{W}{B}^{r} &= \pos{W}{E}^r  - \C{W}{B}  \pos{B}{E_0}, \\ 
    \vel{W}{B}{}^r &=  \vel{W}{E}{}^r - \C{W}{B} \rotvel{B}{}{}^r \times \pos{B}{E_0}.
\end{align}
\end{subequations}
The reference \ac{MAV} orientation $\q{W}{B}^{r}$ is chosen such that the end effector is always perpendicular to the contact surface, assuming perfect position tracking, while the reference force is obtained using the spring model and the nominal displacement of the end effector into the contact surface's normal direction. In our framework each trajectory is accompanied by an appropriate flag which disables or enables the position tracking for the end effector. This is achieved by setting the appropriate gains to zero. In that case the \ac{NMPC} may decide to move the arm to assist the reference tracking of the \ac{MAV} due to the \ac{CoM} displacement. This potentially unwanted behaviour can be avoided by further penalising the arm displacement from its nominal position (i.e.\ by increasing the input gains). However, it is an interesting capability enabled by our hybrid modelling.
\section{Experiments}
\label{sec:experiments}
\subsection{Experimental Setup}
\begin{figure}[htb]
\centering
  \includegraphics[width=0.48\textwidth,trim={0 0 0.0cm 0.0cm},clip]{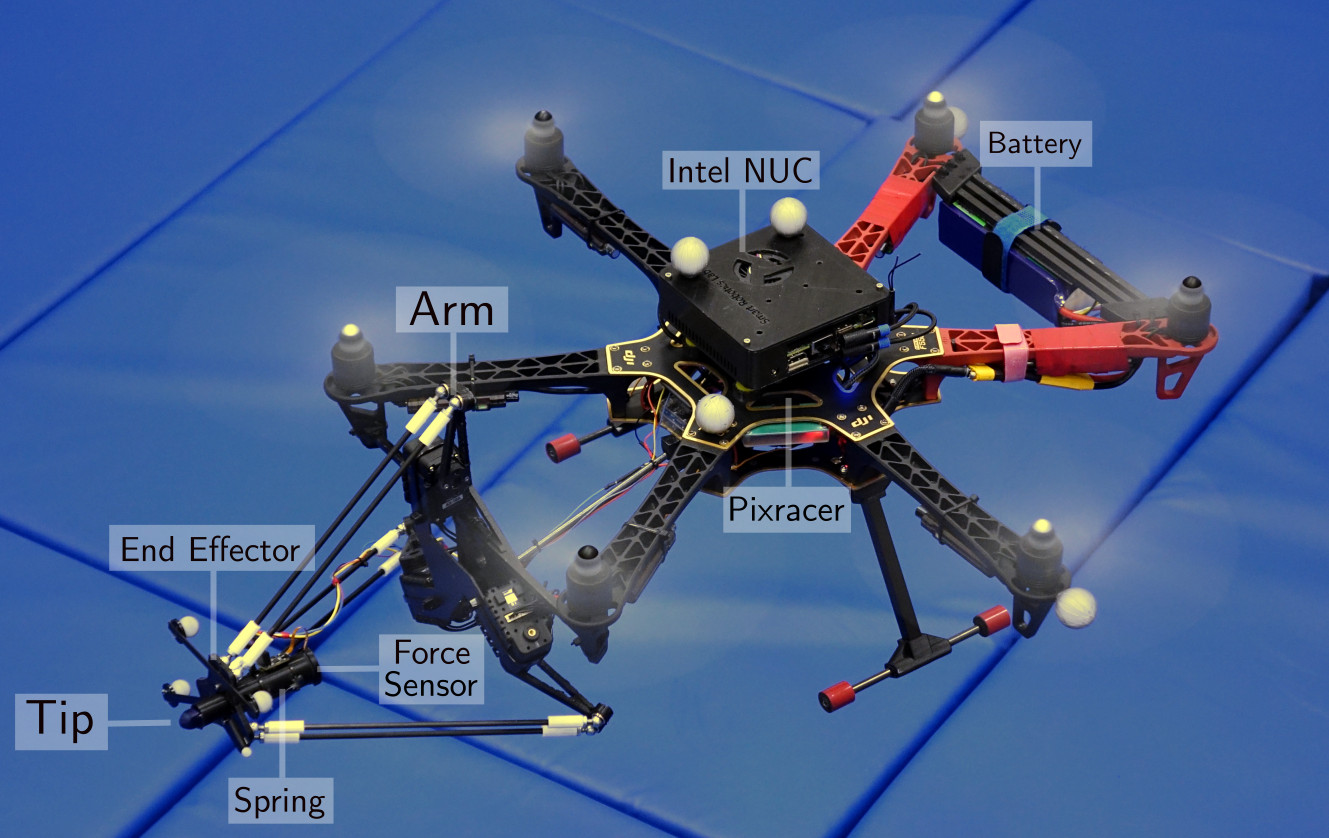}
\caption{The aerial manipulation platform used in the `aerial writing' experiments in Section \ref{sec:experiments} with its individual components labelled.}
\label{fig:drone_w_arm_photo}
\end{figure}

The experiments presented in this section were performed using a custom built hexacopter equipped with a sideways mounted delta arm manipulator. The \ac{MAV} features a frame with a 550~\si{\milli\meter} diameter, a Pixracer flight controller running a modified version of the PX4 firmware, and an Intel NUC-7567U onboard computer running Ubuntu 16.04. It uses 960 KV motors and DJI 9450 propellers. The delta arm uses magnetic universal joints for the connection of the servos with the end effector which maximizes the workspace, minimises backlash and allows the arm to disassemble during possible crashes preventing it from breaking. The arm uses three Dynamixel AX-18A servo motors which are comparably fast and accurate but have limited maximum torque. The system is powered by a 4S 4500 mAh battery and has total weight of 2.6~\si{\kilo\gram}. The end effector holds the pen which is mounted on a spring to provide additional compliance. We set the coefficient of the contact model in \eqref{eq:contact_model} to match the used spring. The applied force is measured by a SingleTact force sensor mounted at the end of the spring. We estimate the spring coefficient $k_s$ by measuring the applied force for known tip displacements. The dimensions of the delta arm were obtained from a highly detailed CAD file and were verified manually. We measured the inertia of the \ac{MAV} $\mathbf{J}$ by measuring its angular response to constant input torque while it is hanging to freely rotate. The thrust to moment coefficient $k_m$ is measured using a thrust stand. A table with the numeric values of the system parameters is given in Table \ref{tab:param} and a photo showing the platform and its different components is shown in Figure \ref{fig:drone_w_arm_photo}.
\begin{table}
\centering
\caption{\label{tab:param}Numeric values of \ac{MAV} and Arm parameters}
\renewcommand{\arraystretch}{1.65}
\begin{tabular}{ll}
    \begin{tabular}{ll}
    \cline{1-2}
    $J_x$ & 0.042~\si{\kilogram\meter\squared}\\
    \cline{1-2}
    $J_y$ & 0.054~\si{\kilogram\meter\squared}\\
    \cline{1-2}
    $J_z$ & 0.110~\si{\kilogram\meter\squared}\\
    \cline{1-2}
    $k_m$ & $1.58\times 10^{-2}$~\si{\newton\meter/\newton}\\
    \cline{1-2}
    $m_e$ & 0.058~\si{\kilo\gram} \\
    \cline{1-2}
    \end{tabular}
&
    \begin{tabular}{ll}
    \cline{1-2}
    $k_s$ & 42.95~\si{\newton/\meter} \\
    \cline{1-2}
    $R$ & 7.2~\si{\centi\meter} \\
    \cline{1-2}
    $r$ & 2.5~\si{\centi\meter} \\
    \cline{1-2}
    $L$ & 6.5~\si{\centi\meter} \\
    \cline{1-2}
    $l$ & 20.2~\si{\centi\meter} \\
    \cline{1-2}
    \end{tabular}
\end{tabular}
\end{table}

We use a Vicon motion capture system to provide external pose estimates. The contact surface is a 
$1\times0.5$~\si{\meter} whiteboard for which we estimate its pose $\T{W}{T}$ based on Vicon measurements. Each experiment consisted of the following different trajectory stages: (i) approach, (ii) write, and (iii) return home. The end effector was enabled, using the appropriate flags as mentioned in Section \ref{sec:trajectory_generation}, for the trajectory writing in (ii) and disabled for the rest. Our analysis mainly focuses on the trajectory writing which includes contact whereas for the other two parts (approach/return) the \ac{MAV} performs simple position tracking. We evaluate the accuracy of our system by comparing the reference trajectories to those estimated by the Vicon motion capture system. In addition, we use a vision-based error as a performance metric. This is because we observed inaccuracies in the Vicon measurements stemming from either bad calibration, poor object visibility, or marker reflections on the whiteboard surface. The visual error is computed by running a 2D \ac{ICP} method \cite{Besl1992} on a filtered and rectified photo of the final writing and a rendering of the planned path. After registration of the two point sets, we use the nearest neighbour distance to evaluate the accuracy.

In the following we show four experiments: in Section \ref{sec:results_trajectory_tracking} we present detailed and repeatable results for two different trajectories, namely \texttt{RSS} and \texttt{$E=mc^2$}. We then show consistent tracking performance across varying \ac{MAV} velocities and text sizes in Sections \ref{sec:results_velocity} and \ref{sec:results_text_size}, respectively.
\subsection{Trajectory Tracking}
\label{sec:results_trajectory_tracking}
\begin{figure*}[htbp]
    \centering
    \includegraphics[width=0.47\textwidth]{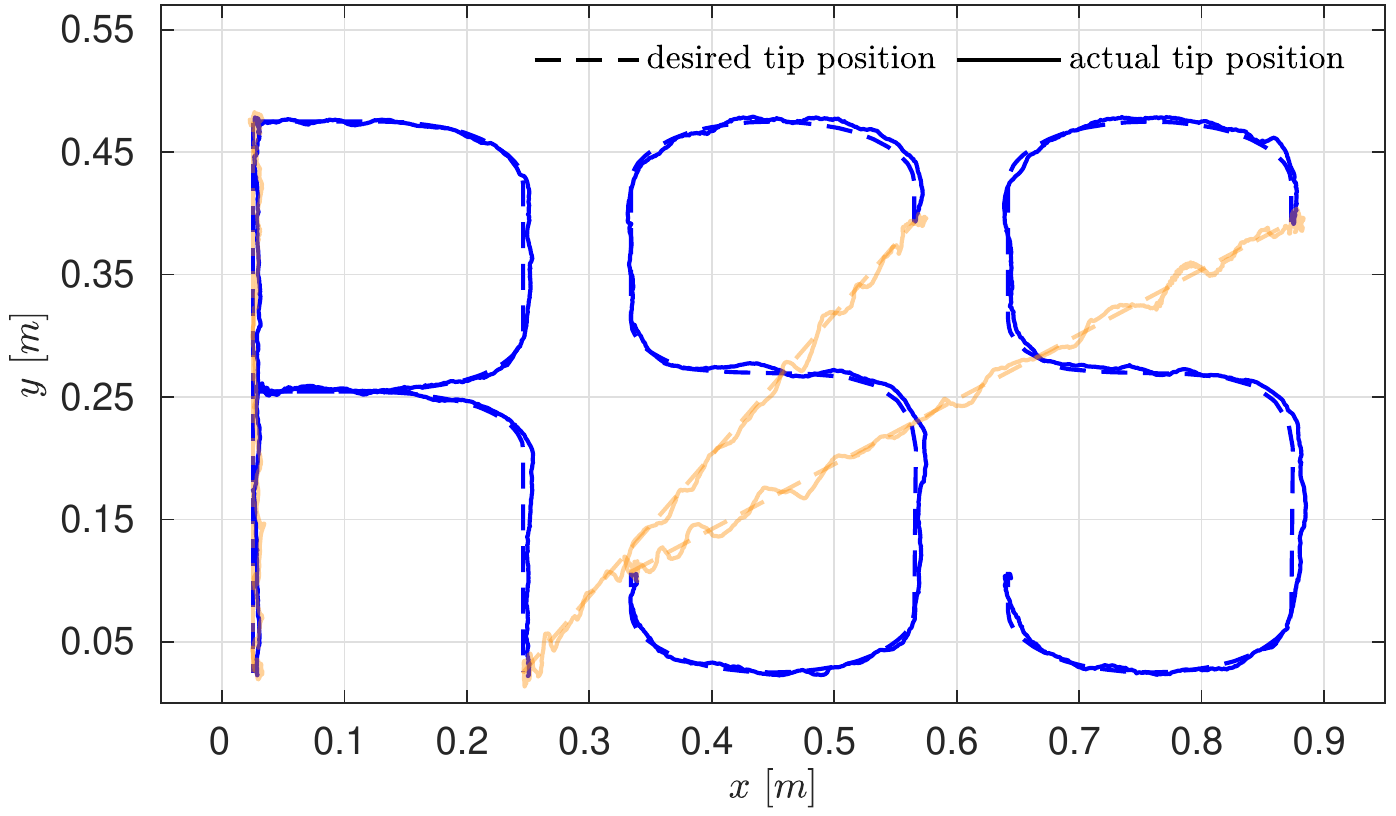}
    \includegraphics[width=0.52\textwidth]{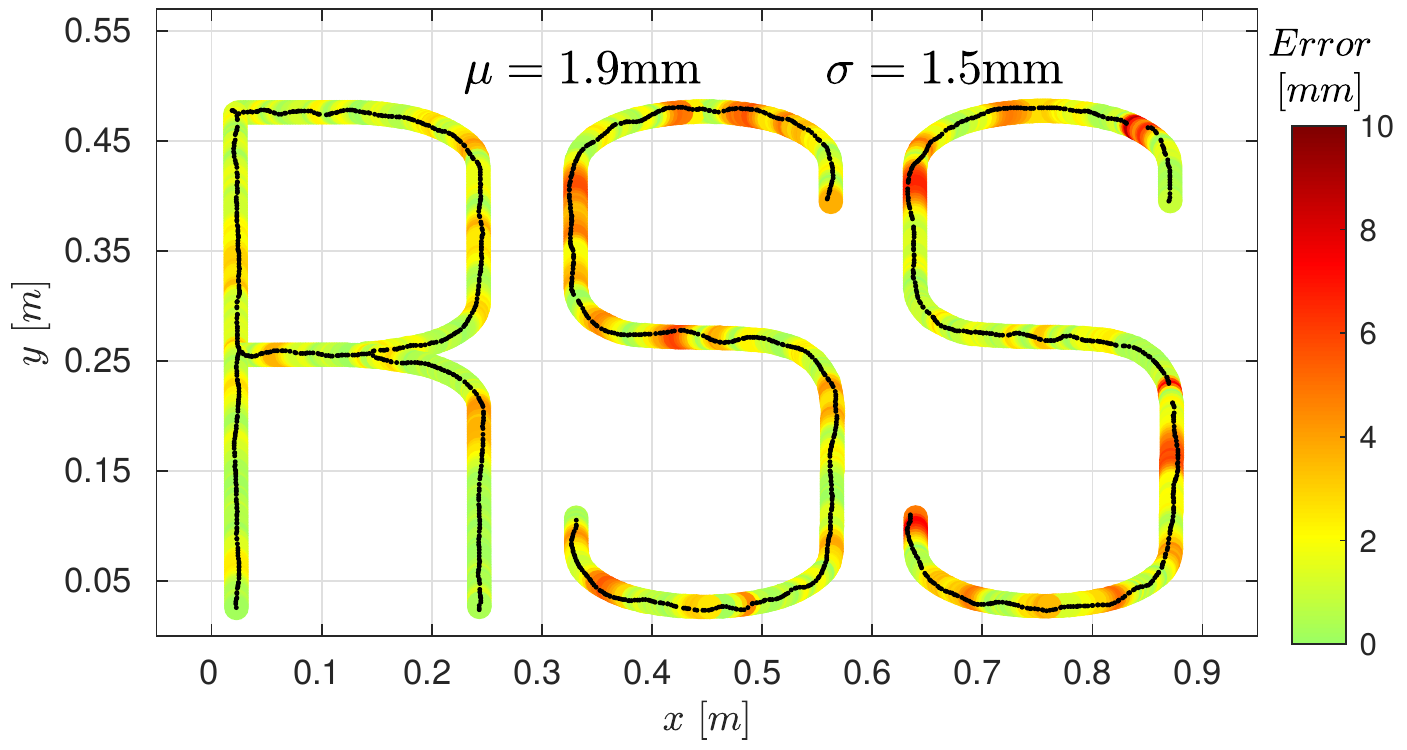}
    \caption{Reference and actual tip position (left) as estimated by Vicon. Blue corresponds to contact  segments while orange refers to free flight. Visual error (right) between reference and actual tip position. The maximum estimated error is lower than 10~\si{\milli\meter} and is located at discontinuous segments as expected.}
    \label{fig:RSS_14_tracking}
    \vspace{7.5pt}
    \includegraphics[width=0.99\textwidth]{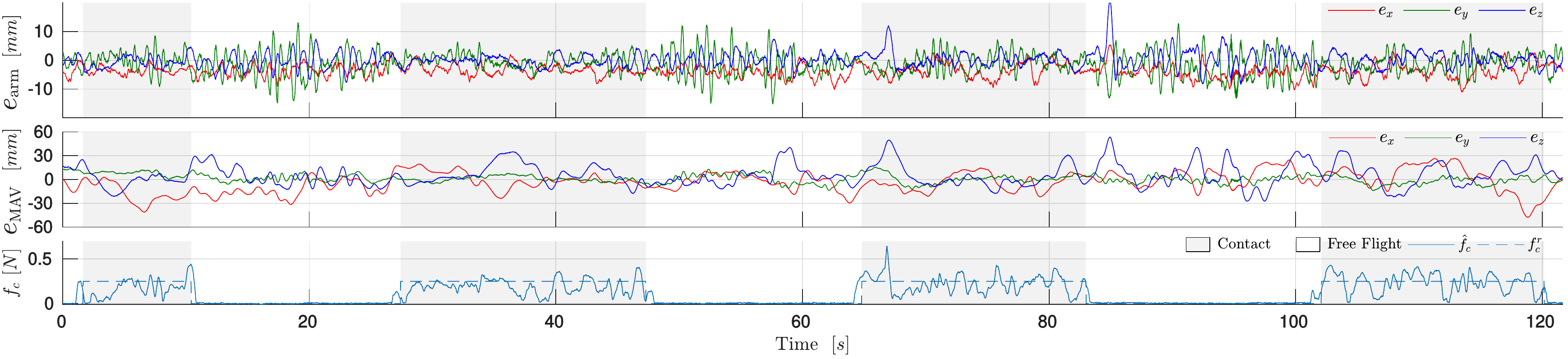}
    \caption{Reference tracking error of the tip position (top), MAV (middle), and measured contact force(bottom). The tracking error is plotted in the contact frame $\cframe{T}$. The tracking accuracy of the end effector is significantly greater than that of the MAV, given that they remain in the $[-10, 10]$~\si{\milli\meter} and $[-40, 40]$~\si{\milli\meter} ranges, respectively.} 
    \label{fig:RSS_14}
\end{figure*}

Figure \ref{fig:RSS_14_tracking} shows the tracking of the \texttt{RSS} trajectory visualised in the contact frame $\cframe{T}$ for the end effector and the MAV. The maximum reference velocity was set to $7.5$~\si{\centi\meter/\second} and the maximum acceleration to $2.5$~\si{\centi\meter/\square\second}. The trajectory consists of four contact segments with a combined duration of 65~\si{\second}. Based on the Vicon estimates the tracking error, shown in Figure \ref{fig:RSS_14}, of the end effector almost always remains in the $[-10, 10]$~\si{\milli\meter} range during the contact segments while the same quantity for the \ac{MAV} is in $[-40, 40]$~\si{\milli\meter} range. This highlights the efficacy of using a manipulator with faster dynamics than the MAV's for precision tasks such as `aerial-writing'. 

\begin{figure*}[htbp]
    \centering
    \includegraphics[width=0.47\textwidth]{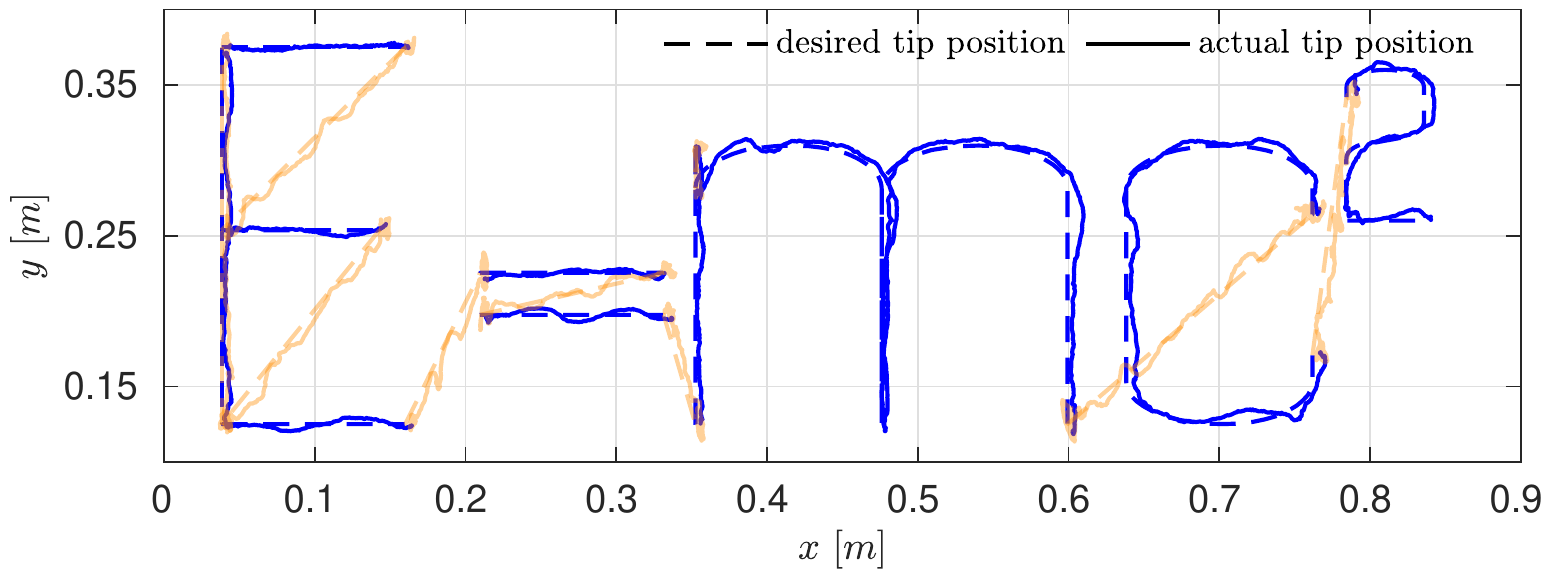}
    \includegraphics[width=0.505\textwidth]{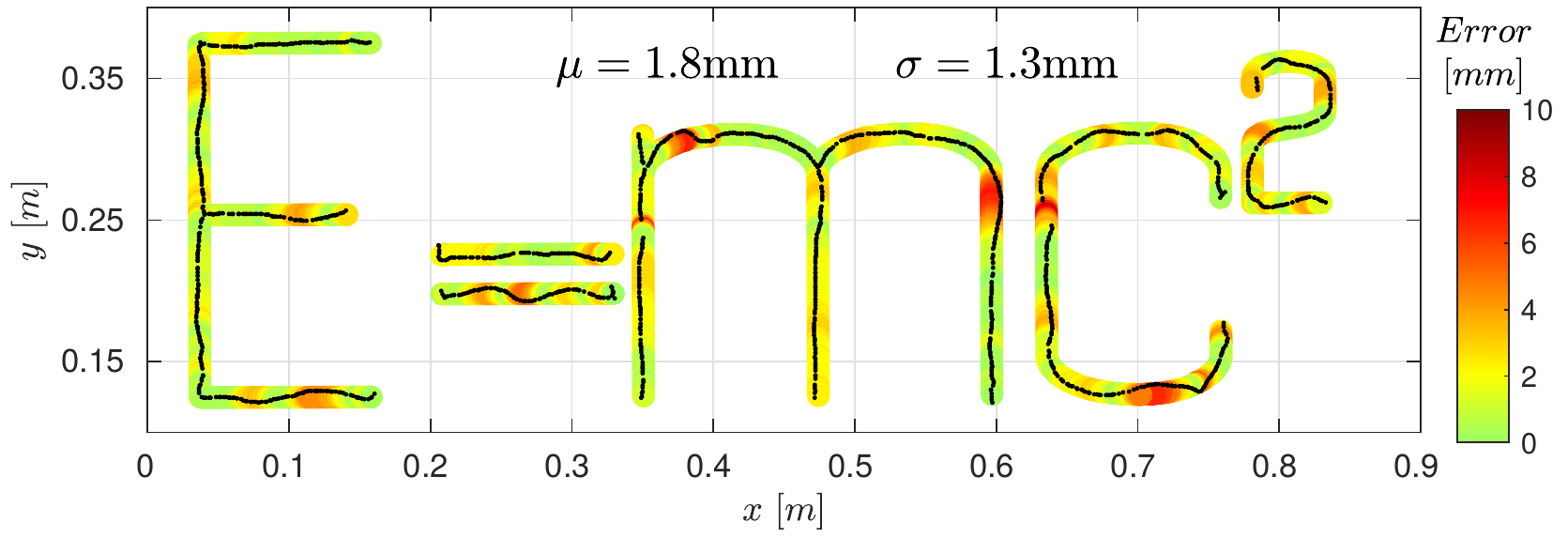}
    \caption{Reference and actual tip position (left) as estimated by Vicon. Blue corresponds to contact segments while orange refers to free flight. Visual error (right) between reference and actual tip position. Similarly as in the \texttt{RSS} experiment shown in Figure \ref{fig:RSS_14_tracking}, maximum error does not exceed 1~\si{\centi\meter}.}
    \vspace{7.5pt}
    \label{fig:EQ_12_tracking}
    \includegraphics[width=0.99\textwidth]{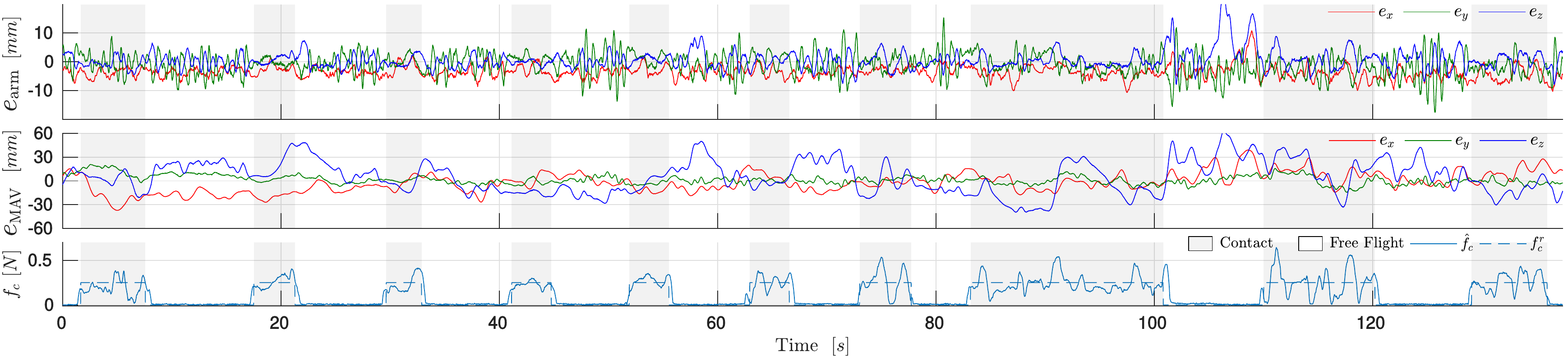}
    \caption{Reference tracking error of the tip position (top), MAV (middle) and measured contact force (bottom). The tracking accuracy of the end effector is significantly greater than that of the MAV, given that they remain in the $[-10, 10]$~\si{\milli\meter} and $[-50, 50]$~\si{\milli\meter} ranges, respectively}
    \label{fig:EQ_12}
\end{figure*}

Similarly to the above, Figure \ref{fig:EQ_12_tracking} shows the trajectory tracking for the more challenging \texttt{$E=mc^2$} experiment which contains ten contact segments with a combined duration of 63~\si{\second}. Tracking accuracy is similar as before with the end effector and MAV tracking error in the  $[-10, 10]$ and $[-50, 50]$~\si{\milli\meter} range. The accuracy can be visually verified since the overlapping segments of the `R' and `m' coincide almost perfectly. Additionally, the consistent approaching and retracting from the contact surface leads to identical starting points of individual letter segments, e.g.\ the three horizontal lines of the letter `E'.
In both cases, the maximum error based on the visual error analysis is 10~\si{\milli\meter} mostly originating from temporary loss of contact. Possible reasons for this are bad estimation of the orientation part of the contact frame transformation $\T{W}{T}$, the assumption of a perfectly flat contact surface being incorrect but also the finite accuracy of the delta arm. The imperfect tracking along the contact frame normal direction (shown in blue in Figures \ref{fig:RSS_14}, \ref{fig:EQ_12}) is also reflected in the reference force tracking. 

\begin{figure*}[htbp]
    \centering
    \includegraphics[width=0.47\textwidth]{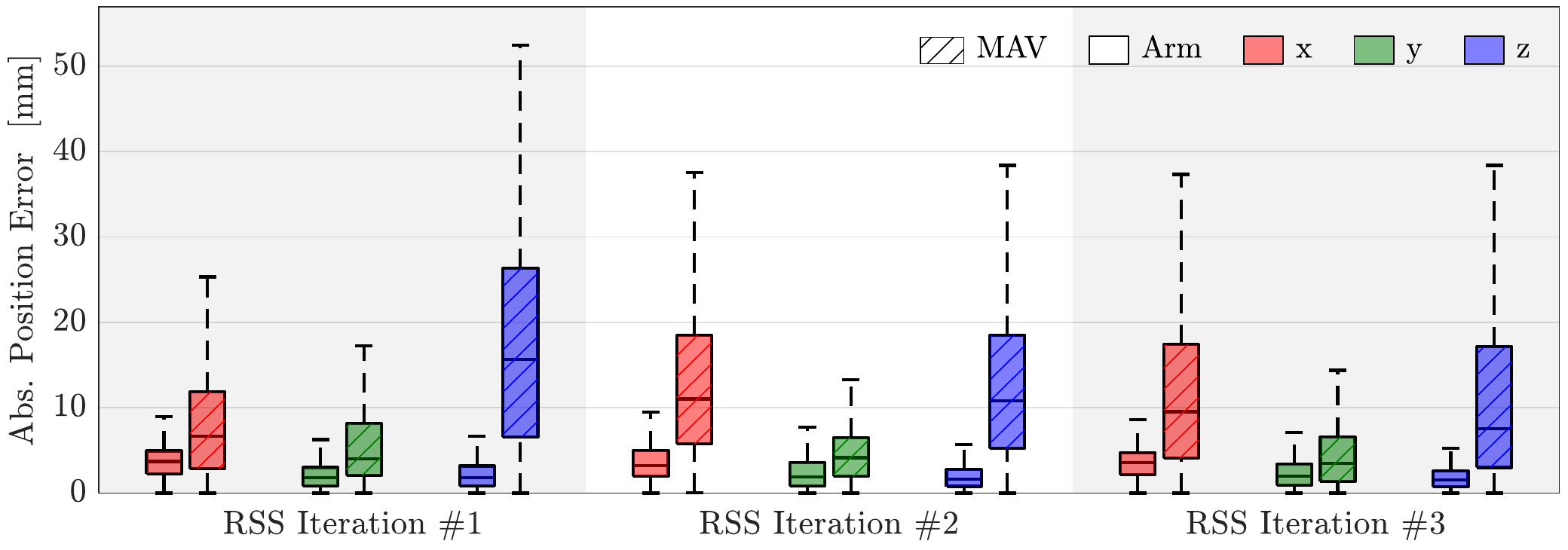}
    \includegraphics[width=0.47\textwidth]{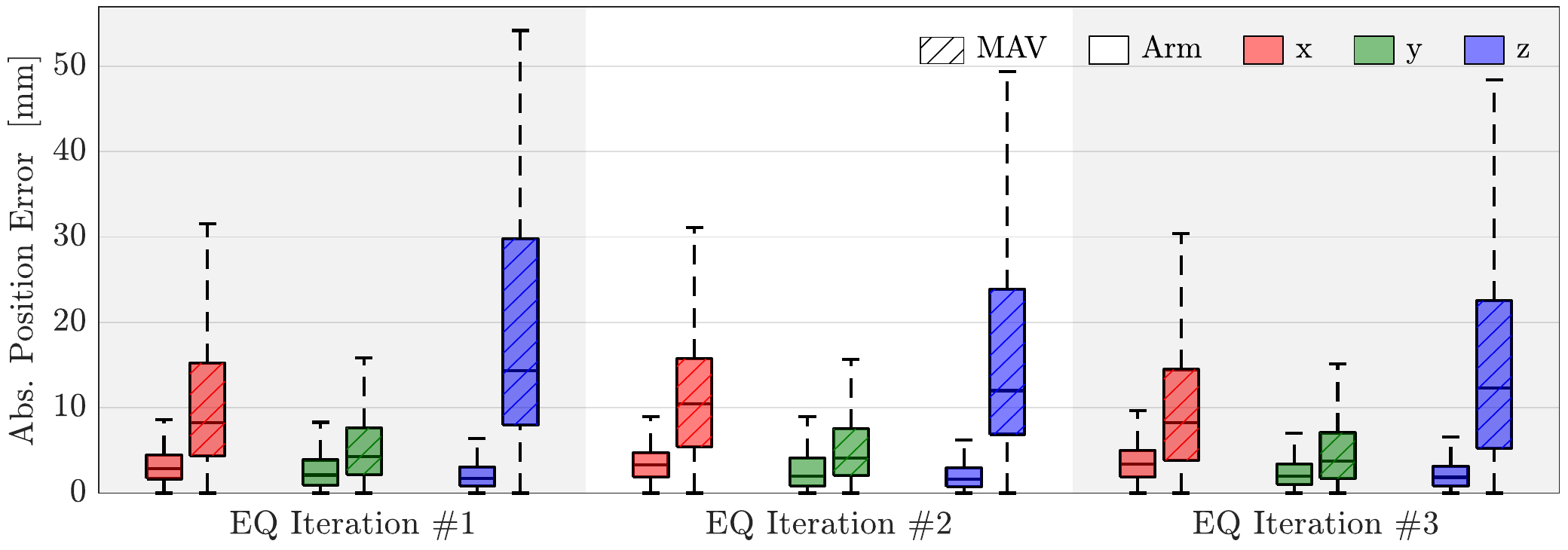}
    \caption{MAV and end effector box plots of the contact segments for 3 iterations of the \texttt{RSS} trajectory experiment (left) and the more challenging \texttt{$E=mc^2$} trajectory experiment (right).}
    \label{fig:repeat_experiments}
\end{figure*}

To prove the repeatability of our approach, we conducted each experiment thrice. We give the relevant tracking statistics for the MAV and arm separately in Figure \ref{fig:repeat_experiments}, in which the textured box plots correspond to MAV data and the plain ones to that of the end effector. The median and extreme values for the end effector are significantly lower than the ones for the MAV and consistent with the values reported above. This further shows the benefit of using an aerial manipulator for precise tasks including contact. The MAV tracking accuracy along the $z$ axis was the lowest amongst all axes, as this was most affected by the interaction forces and unmodelled torque disturbances due to the movement of the servos.

\subsection{Velocity Sweep}
\label{sec:results_velocity}
The aim of the next experiments is to demonstrate the effects of the input velocity and acceleration on the writing accuracy. We performed five iterations of the same \texttt{Hello} trajectory experiment with different velocity and acceleration profiles. 
The different iterations correspond to maximum velocity $v_{max} \in \{7.5, 12.5, 17.5, 22.5, 27.5\}$~\si{\centi\meter/\second} and maximum acceleration $a_{max} \in \{3.75, 6.25, 8.75, 11.25, 13.75\}$~\si{\centi\meter/\square\second}.

\begin{figure*}[htbp]
    \centering
    \includegraphics[width=0.99\textwidth]{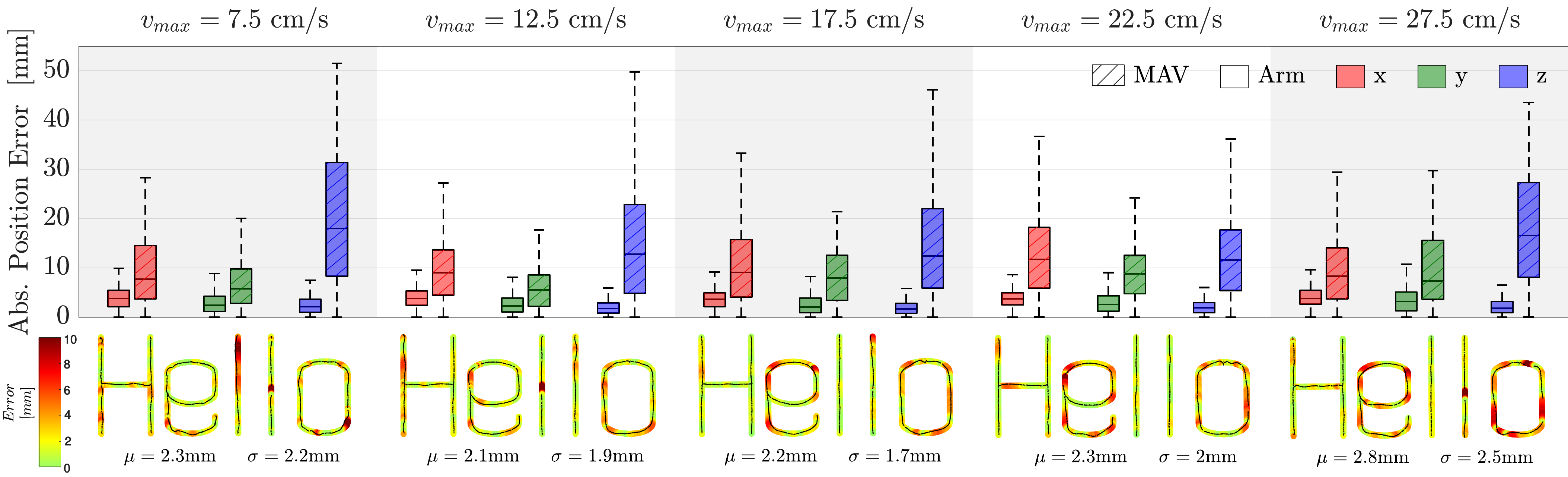}
    \caption{MAV and end effector box plots (top) and visual errors (bottom) for 5 iterations of the \texttt{Hello} trajectory. Different iterations correspond to different velocity and acceleration profiles.}
    \label{fig:varying_velocity_experiments}
\end{figure*}

Figure \ref{fig:varying_velocity_experiments} shows the box plots for the MAV and end effector tracking accuracy based on the Vicon measurements. The plots show consistent tracking results in all the different velocity and acceleration settings tested. 
The numeric values of the tracking error are similar to the ones presented in Section \ref{sec:results_trajectory_tracking} with the end effector achieving sub centimetre accuracy (per axis) while the MAV error is consistently less than $50$~\si{\milli\meter}. However, by observing the visual error, we see that as the reference velocity increases, the system struggles more with the segments containing curvature e.g.\ `e' and `o'. In contrast, the performance on the straight line segments remains similar.

We believe that the tracking error of the MAV can be further reduced by giving the \ac{NMPC} dynamically feasible trajectories not only for position and velocity but also acceleration, jerk, and snap. Regarding the end effector tracking error, we generally expect this to increase for reference velocities beyond the ones tested here. This is due to the mismatch between the formulated control model and the real one. 

\subsection{Text Size Sweep}
\label{sec:results_text_size}
\begin{figure}[htbp]
    \centering
    \includegraphics[width=0.4\textwidth]{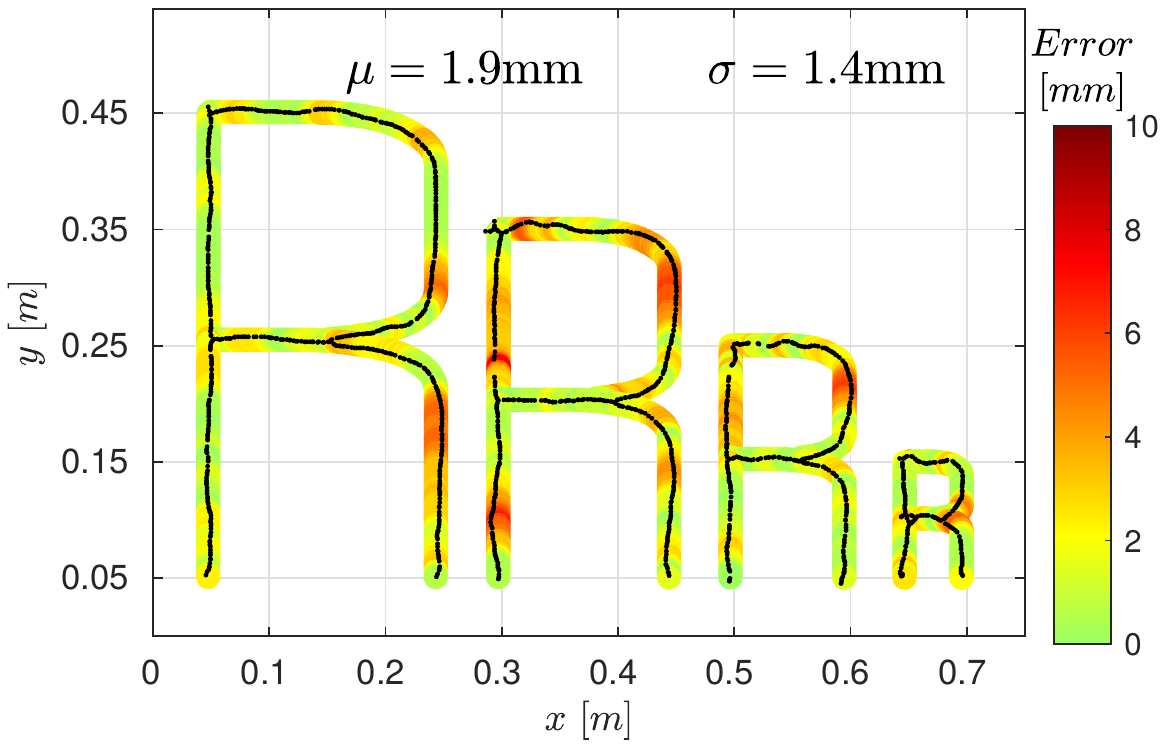}
    \vspace{-1.0em}
    \caption{Visual error plot showing consistent results for varying text sizes.}
    \label{fig:text_size_experiments}
\end{figure}

In Figure \ref{fig:text_size_experiments} we show the visual error of our system for the same trajectory in four different text sizes ranging from 10 to 40~\si{\centi\meter}. The consistent accuracy shows that the system can handle the fast direction changes imposed by the small scale.  
\section{Discussion}
\label{sec:discussion}
Overall, our system achieves accurate and consistent results over a series of different trajectories. The tracking error of the end effector is significantly lower than that of the MAV; highlighting the accuracy boost due to the utilisation of the arm. We would like to mention that our system was built using relatively low-cost off-the-shelf components and 3D printed parts. This leads to errors in the manufacturing with respect to the reference model, e.g.\ errors in the true inverse kinematics of the arm due to non-identical dimensions of its links.

Another important issue that we faced during our experiments was the reliance on the motion capture system for localisation. Apart from issues related to WiFi delays, which resulted in temporary loss of tracking, we faced problems with poor object visibility resulting in unreliable estimates of both the static objects, such as the contact frame, and moving ones such as the MAV. In fact, during our data analysis we realised that there are segments where Vicon returned mechanically impossible configurations for our system e.g.\ end effector position below the surface of the contact frame. Despite these problems which further propagate into tracking errors, our system successfully handled multiple transitions to contact during the same experiment. 

We experimentally verified that for contact tasks, where the main objective is accuracy instead of speed, using a planner respecting full state dynamic feasibility is not an absolute necessity. Despite our simplified motion planner, our system achieves sub-centimetre accuracy. However, we argue that for more aggressive maneuvers, a full state dynamically feasible planner would be required.
\section{Conclusion}
\label{sec:conclusion}
We have presented a hybrid model-based algorithm for aerial manipulation using an underactuated \ac{MAV} with an attached end effector performing `aerial-writing' tasks on a whiteboard. We demonstrated our system in a series of  experiments with trajectories requiring multiple transitions from free-flight to contact and vice-versa. The end effector tracking error consistently remains in the $[-10, 10]$~\si{\milli\meter} range for trajectories with maximum velocities ranging from 7.5 to 27.5~\si{\centi\meter/\second}, maximum acceleration from 3.75 to 13.75~\si{\centi\meter/\square\second} and text sizes from 10 to 40~\si{\centi\meter}. All algorithms run in real-time onboard the MAV. We believe that our method is generic enough to be applied to different types of \acp{MAV} and manipulators. We further believe that our framework can be extended to more practical applications such as physical interaction with surfaces e.g.\ drilling, welding, grinding or inspection through contact.

Regarding future work, online estimating $\T{W}{T}$ and closing the manipulation task loop with visual feedback would constitute a major improvement to correct for errors stemming from bad calibration, Vicon delays and inaccuracies of the arm. This will allow us to better handle the system's sensitivity to imperfect estimates of transformations such as the contact frame $\T{W}{T}$ and arm to body transformation $\T{B}{A}$. The current method further assumes an instantaneous end effector position response, which we plan to substitute with a more realistic model or ultimately formulate the full multi-body dynamics model which would result in a performance boost mainly for faster maneuvers.  Finally, we aim to implement and test our method using omnidirectional \acp{MAV} which due to their ability of generating lateral forces have superior force exertion capabilities compared to the one presented in this work.
\bibliographystyle{plainnat} 
\bibliography{references/2020-rss-tzoumanikas.bib}
\end{document}